\documentclass[final]{cvpr}

\usepackage{times}
\usepackage{epsfig}
\usepackage{graphicx}
\usepackage{amsmath}
\usepackage{amssymb}

\usepackage[pagebackref=true,breaklinks=true,colorlinks,bookmarks=false]{hyperref}

\def\cvprPaperID{27} 
\def\confYear{CVPR 2021}

\begin{document}

\title{Florida Wildlife Camera Trap Dataset}

\author{Crystal Gagne \and Jyoti Kini \and Daniel Smith \and Mubarak Shah \and
University of Central Florida, Orlando, Florida \\
{\tt\small \{crystal.gagne, jyoti.kini\}@knights.ucf.edu, daniel.smith@ucf.edu, shah@crcv.ucf.edu}
}
\maketitle

\begin{abstract}
   Trail camera imagery has increasingly gained popularity amongst biologists for conservation and ecological research. Minimal human interference required to operate camera traps allows capturing unbiased species activities. Several studies - based on human and wildlife interactions, migratory patterns of various species, risk of extinction in endangered populations - are limited by the lack of rich data and the time-consuming nature of manually annotating trail camera imagery. We introduce a challenging wildlife camera trap classification dataset collected from two different locations in Southwestern Florida, consisting of 104,495 images featuring visually similar species, varying illumination conditions, skewed class distribution, and including samples of endangered species, i.e. Florida panthers. Experimental evaluations with ResNet-50 architecture indicate that this image classification-based dataset can further push the advancements in wildlife statistical modeling. We will make the dataset publicly available.  
\end{abstract}

\section{Introduction}
Most conservation biologists recognize the utilitarian value of biodiversity for human existence \cite{cardinale2012biodiversity, newbold2015global}. Therefore, numerous research studies have focused on protecting and restoring wildlife by monitoring their population distribution, identifying threats to their survival, and eliminating the perils. Our collaboration primarily focused on the conservation of the endangered Florida panther (Puma concolor coryi), which is one of the most endangered species of mammals in the United States, with less than 230 individual adults estimated to be living in South Florida \cite{mcclintock2015endangered}. \\

Measuring abundance is central to endangered species recovery and essential for management, but accurate estimates of population size are difficult to obtain for rare species using traditional field methods. While researchers have relied on field studies involving direct observation, mark-recapture, and radio-telemetry to monitor species population and their behavior in the past; technological advancements and ease of access to non-invasive inexpensive trail cameras have drawn the attention of present-day investigators. Trail camera imagery has a wide variety of biological and ecological applications in conservation \cite{o2010camera, yu2013automated, miguel2016finding, swanson2015snapshot}. For our project, two different locations in South Florida, namely Corkscrew Swamp (Corkscrew) and Okaloacoochee Slough State Forest (OKSSF), were surveyed between January 2018 to late 2019. Each trail camera took up to 40,000 images during the first couple of months, and by the end of the survey, over three-quarters of a million images were generated. The images captured using these cameras, first, had to be curated for efficient information retrieval. It took many volunteers several months to manually sort and organize the data into file systems, eventually creating a clean dataset of more than 100k images and 22 classes, including the endangered Florida panther. \\

Wildlife populations can be monitored via camera trap studies, yet processing the data generated in these studies is an enormous task. Due to the time-consuming nature of manually annotating camera trap images, scientists have begun to experiment with different methods of automatically classifying images using deep neural networks. There is a growing literature base demonstrating the power of AI in recognizing images of wildlife \cite{willi2019identifying, swanson2015snapshot, gomez2016animal, villa2017towards, norouzzadeh2018automatically, tabak2019machine}. Recent studies have demonstrated that deep neural networks can automatically identify images of different wildlife species with accuracy on par of human volunteers, saving over 8.4 years at 40 hours per week of volunteer labeling effort \cite{norouzzadeh2018automatically, schneider2019past}. AI can dramatically improve the speed and efficiency of species identification and data management. \\

Contemporary models used to classify images from trail cameras, however, do not perform well when used on images from a different location compared to the training data, which implies that the models are non-transferable \cite{tabak2020improving}. Model transferability is vital if the model is to be used by biologists for more than one study area. We introduce a challenging real-world condition-based wildlife dataset, which will promote the need for better generalization, thereby promoting advances in this area. Using a ResNet-50 architecture, we train a preliminary transferable model on images of the Florida panther and other species available in the dataset. 

\begin{figure*}
\begin{center}
    \includegraphics[width=0.75\textwidth,height=0.45\textheight]{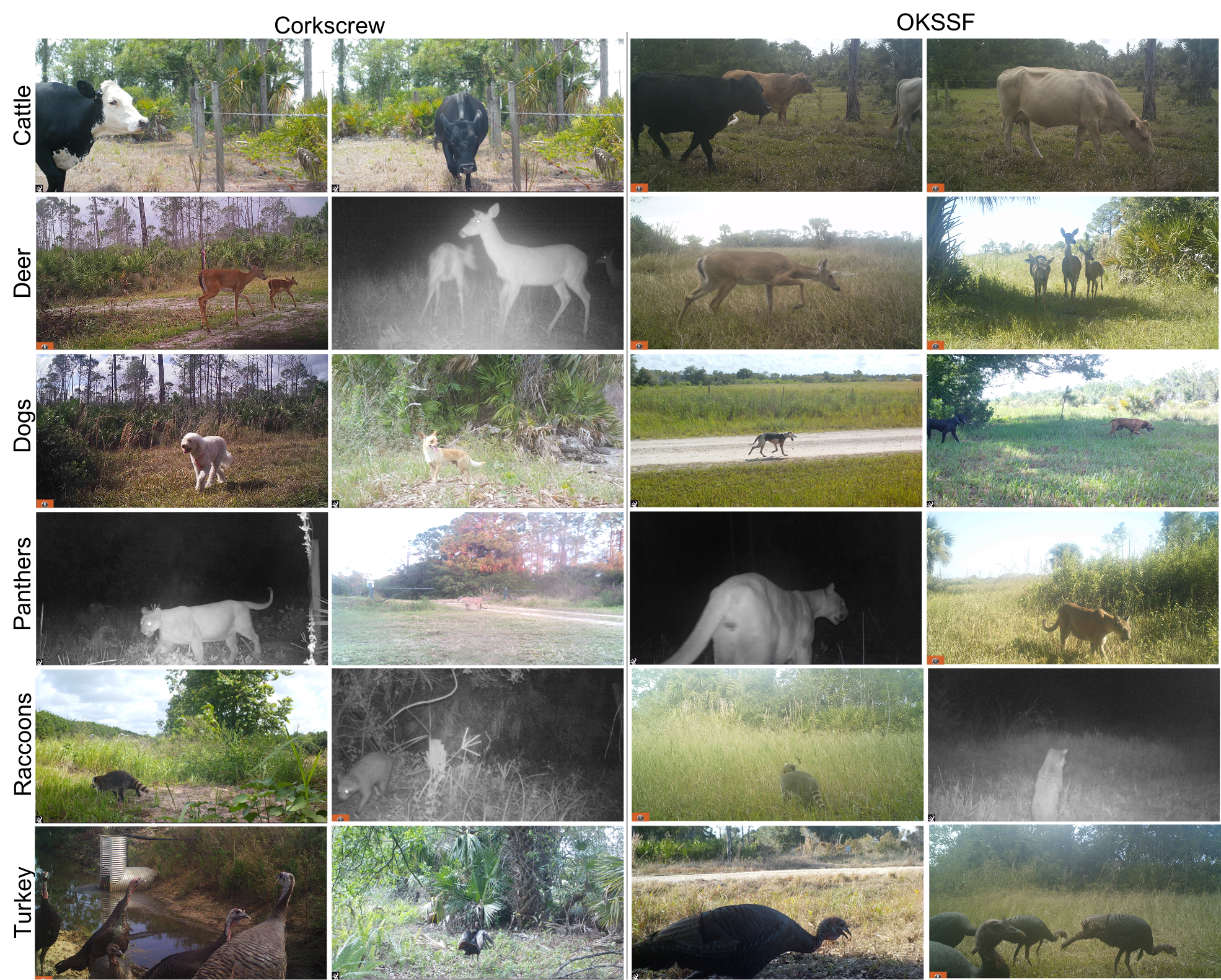}
\end{center}
   \caption{Examples of images from the dataset. The first two columns exhibit samples from the Corkscrew location, while the next set of columns contain specimens from the OKSSF location.}
\label{fig:Examples}
\end{figure*}

\section{Dataset Overview}
\subsection{Trail Camera Images}
Our dataset is composed of 104,495 images from both of our trail camera study locations combined (see Table \ref{table:Dataset}). The dataset contains over 2500 images of Florida panthers. Due to the rarity of this species, this dataset has the potential to be invaluable as a conservation tool for this species. The dataset has images with varying backgrounds taken from several sites at the two primary locations surveyed. The locations of survey were chosen due to their proximity to major roads and the relatively high level of panther activity in these areas. Sample images from the dataset can be viewed in Figure \ref{fig:Examples}.

\begin{figure}[t]
\begin{center}
   \fbox{\includegraphics[width=1\linewidth]{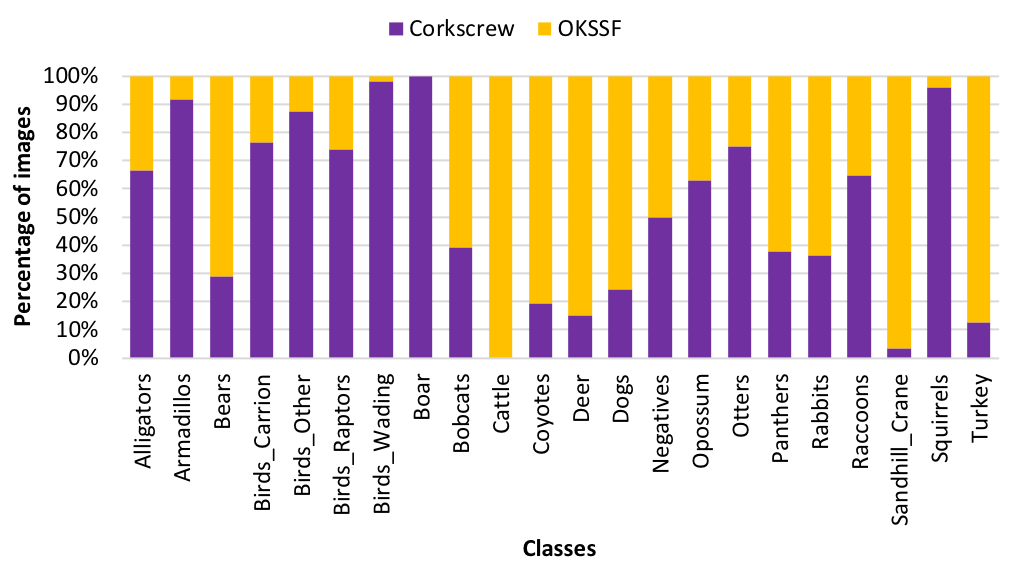}}
\end{center}
   \caption{Location-based distribution of images per species.}
\label{fig:long}
\label{fig:Distribution}
\end{figure}

\setlength{\tabcolsep}{3.9pt}  
\begin{table}
\renewcommand{\arraystretch}{1.06}
\begin{center}
\begin{tabular}{|l|c|c|c|c|}
\hline
{\bf {Classes}} &
        {\bf {Corkscrew}} & 
        {\bf {OKSSF}} &
        {\bf {Per-Class}} &
        {\bf {\% of}} \\
        &
        & 
        &
        {\bf {Total}} &
        {\bf {Overall}} \\ \hline \hline
        Alligators      &   18	    &   9	    &   27      &   0.03 \\ \hline
        Armadillos      &	119	    &   11      &   130     &   0.12 \\ \hline
        Bears           &	149	    &   363     &	512     &	0.49 \\ \hline
        Birds\_Carrion	&   1028    &	318     &	1346    &	1.29 \\ \hline
        Birds\_Other     &	2146    &	304     &	2450    &	2.34 \\ \hline
        Birds\_Raptors   &	54      &	19      &	73      &	0.07 \\ \hline
        Birds\_Wading    &	3741    &	80	    &   3821    &	3.66 \\ \hline
        Boar            &	117 	&   0       &	117     &	0.11 \\ \hline
        Bobcats         &	966     &	1497    &	2463    &	2.36 \\ \hline
        Cattle          &	5       &	22297   &	22302   &	21.34 \\\hline
        Coyotes	        &   28      &	118     &	146	    &   0.14 \\ \hline
        Deer	        &   5415    &	30780   &	36195	&   34.64 \\\hline
        Dogs            &	48      &	148     &	196     &	0.19 \\ \hline
        Negatives	    &   8999	&   9000    &	17999   &	17.22 \\\hline
        Opossum         &	882     &	519     &	1401    &	1.34 \\ \hline
        Otters	        &   18      &	6       &	24      &	0.02 \\ \hline
        Panthers        &	965     &	1593    &	2558    &	2.45 \\ \hline
        Rabbits         &	589     &	1031    &	1620    &	1.55 \\ \hline
        Raccoons        &	661     &	359     &	1020    &	0.98 \\ \hline
        Sandhill\_Crane  &	39      &	1120    &	1159    &	1.11 \\ \hline
        Squirrels       &	501     &	21      &	522     &	0.50 \\ \hline
        Turkey          &	1070    &	7344    &	8414    &	8.05 \\ \hline
        \hline
        {\bf {Overall}} &
        {\bf {27558}}   &
        {\bf {76937}}   &	
        {\bf {104495}}  &	
        {\bf {100}} \\
        \hline
\end{tabular}
\end{center}
\caption{A comprehensive list providing the number of images collected per species from two different locations in our dataset.}
\label{table:Dataset}
\end{table}

\subsection{Data  Challenges}
\begin{figure}[t]
\begin{center}
   \includegraphics[width=1\linewidth]{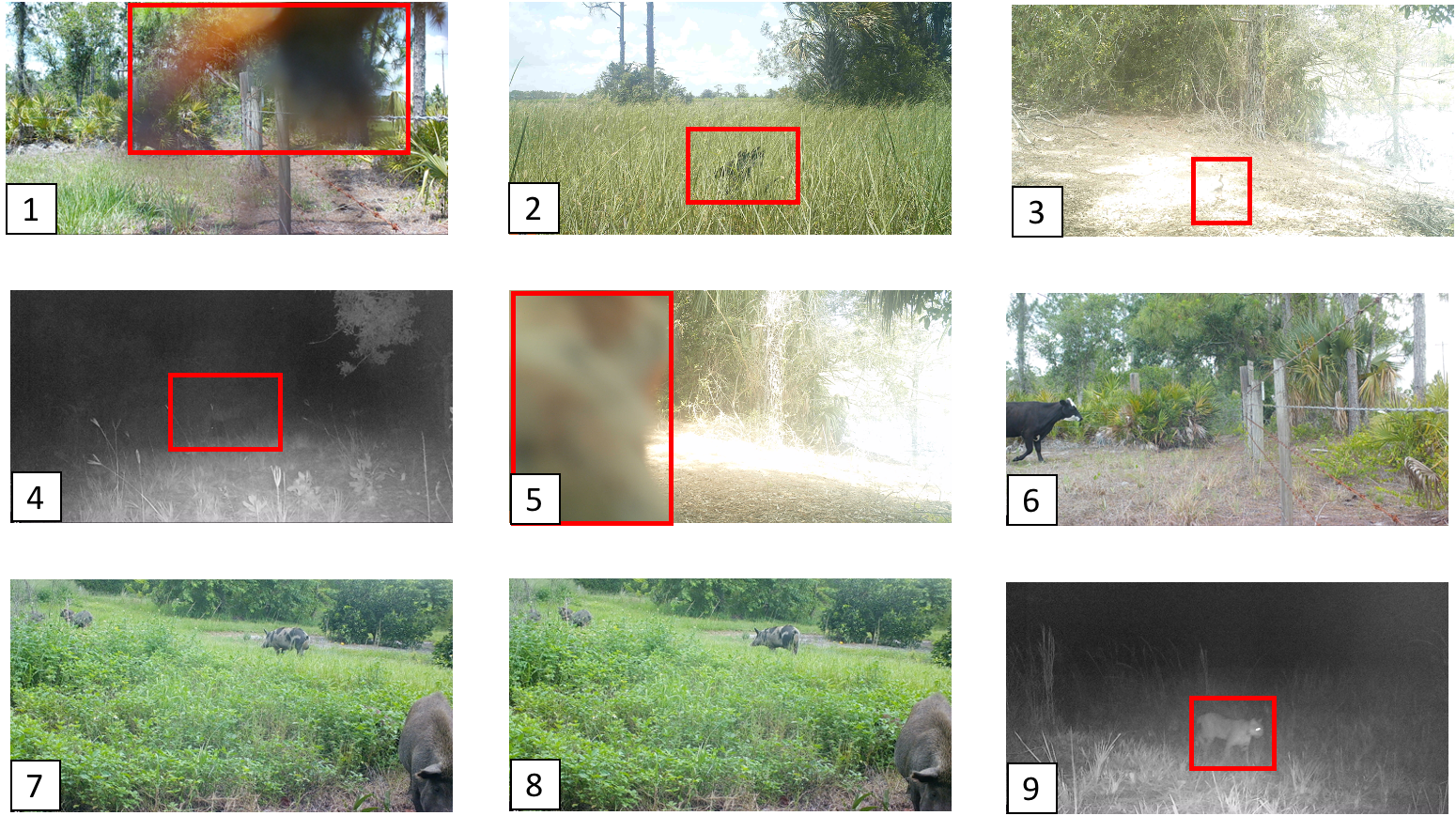}
\end{center}
   \caption{Specimen images depicting the highlighted challenges associated with the dataset: (1) Motion blur, (2) Occlusion, (3) Camouflage,  (4) Illumination issues, (5) Excessive proximity of the animals to the camera traps making it impossible to identify salient features, (6) Imbalance in the dataset due to substantial difference in the numbers of images for classes such as cattle/deer, (7/8)  Multiple images with hardly any difference captured by continuous camera triggers, and (9) View-point based deception resulting in bobcats appearing similar to panthers.}
\label{fig:Challenges}
\end{figure}

Similar to other trail camera datasets, such as Beery et al.’s iWildCam 2018 Challenge Dataset \cite{beery2019iwildcam}, our dataset exhibits a range of inherent challenges, as illustrated in Figure \ref{fig:Challenges}. Since the trail camera captures images throughout the day, multiple samples suffer from illumination variations (either over or under saturated).  Frequently, images turn blurry when the subject moves too quickly, and several times blur could be present from moisture on the lens. In the case of some cameras, IR system malfunction has resulted in “pink-washed” daytime photos. Additionally, the distance between animal and camera does not remain consistent; hence scale, focus, and visibility vary considerably even in subsequent images. Sometimes, animals tend to be almost entirely obscured by vegetation in the background, or they could be so close to the camera only an indiscernible part of their body would be visible. In addition, the image backgrounds are non-static, and vegetation could shift quite a lot from one image to the next. While all these complexities add real-world challenges to the dataset, these limitations occasionally make it impossible for the image to be identified. Such unidentifiable images have been omitted from the dataset. There is also a margin of error for what has been sorted, around 2\%. 

\section{Baseline Experiments}
\begin{figure}[t]
\begin{center}
   \includegraphics[width=1\linewidth]{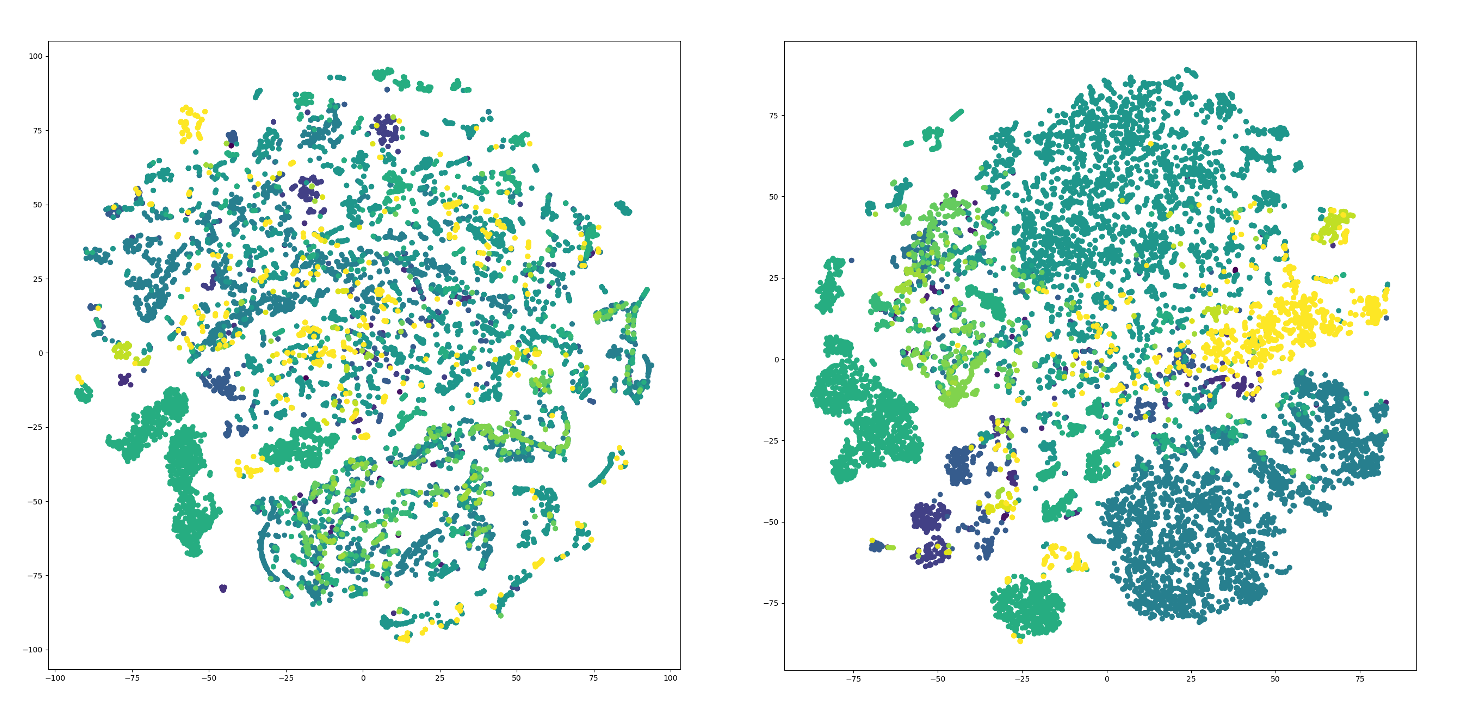}
\end{center}
   \caption{t-SNE visualization of randomly initialized features (left) and learned embeddings (right) for test images in the dataset.}
\label{fig:TSNE}
\end{figure}

We train a baseline model using ResNet-50 architecture. The dataset comprising 104,495 samples is split into train, validation, and test set in 70\%-10\%-20\% proportion. We have also employed random cropping, horizontal flip, and rotation for data augmentation. Images resized to $224\times224$px are processed through the model in a batch size of 64, using a learning rate of 1e-3. We achieve a baseline test accuracy of 78.75\% on our ResNet-based model. For the test images, we do a t-SNE visualization of the output embeddings, as shown in Figure \ref{fig:TSNE}, to demonstrate the effectiveness of our method.

\begin{figure}[t]
\begin{center}
   \fbox{\includegraphics[width=1\linewidth]{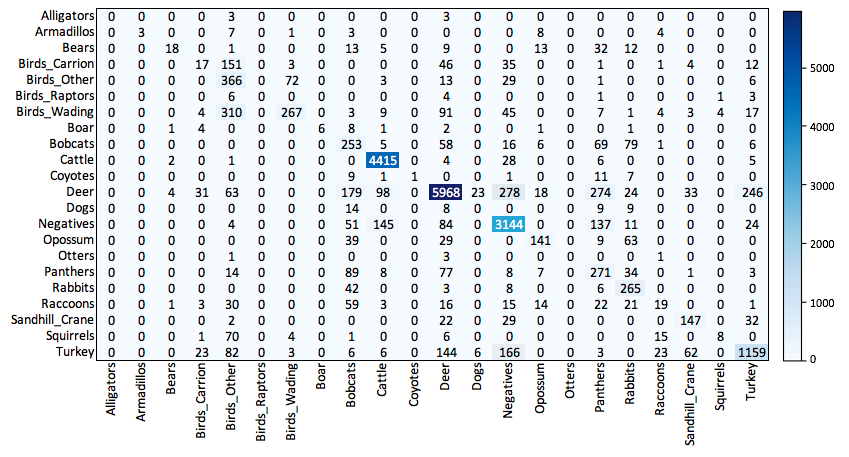}}
\end{center}
   \caption{Confusion matrices illustrating results on the test data for 22 classes.}
\label{fig:Confusion}
\end{figure}

\section{Conclusions and Future Work}
In this paper, we have introduced a camera trap dataset comprising 100k+ wildlife samples belonging to 22 different categories. The complexity of the landscape, adverse lighting conditions, unconventional subject behavior captured in the dataset, using trail cameras, has aimed to improve the ability of automated systems to generalize to new environments. While the dataset provides a good platform to work on domain adaptation and data imbalance issues, we also have some additional unlabeled data which will allow us to explore the area of semi-supervised learning. Although this project originally focused only on monitoring the ecosystem concerning the endangered Florida panther, the curated trail camera images have presented avenues to explore several areas of biodiversity conservation. The dataset will be released publicly to enable access to a broader community and foster development in biodiversity research. 

{\small
\bibliographystyle{ieee_fullname}
\bibliography{egbib}
}

\end{document}